\documentclass{article}

\usepackage{arxiv}

\usepackage[utf8]{inputenc} 
\usepackage[T1]{fontenc}    
\usepackage{hyperref}       
\usepackage{url}            
\usepackage{booktabs}       
\usepackage{amsfonts}       
\usepackage{nicefrac}       
\usepackage{enumitem}

\usepackage{nicematrix}
\usepackage{threeparttable}

\usepackage{microtype}      
\usepackage{lipsum}		
\usepackage{graphicx}
\usepackage{natbib}
\usepackage{doi}
\usepackage{amsmath}
\usepackage{float}
\usepackage{pgfplots}
\usepackage{subcaption}
\usepackage[normalem]{ulem}

\title{Parameter-Efficient Checkpoint Merging via Metrics-Weighted Averaging}

\author{Shi Jie Yu \\
        Tensorplex Labs\\
        \texttt{shijie@tensorplex.ai}
        \And
	Sehyun Choi\thanks{Work done while at Tensorplex Labs.} \\
	Twelve Labs \\
	\texttt{sam.choi@twelvelabs.io}
}

\date{}

\renewcommand{\shorttitle}{Metrics-Weighted Averaging}

\pgfmathsetmacro{\calcDiffMathA}{0.22826 - 0.21728}
\pgfmathsetmacro{\calcDiffMathB}{0.22617 - 0.21728}
\pgfmathsetmacro{\calcDiffMathC}{0.22584 - 0.21728}
\pgfmathsetmacro{\calcDiffMathD}{0.2248 - 0.21728}
\pgfmathsetmacro{\calcDiffMathE}{0.22471 - 0.21728}

\pgfmathsetmacro{\calcDiffSafetyA}{0.4920 - 0.4872}
\pgfmathsetmacro{\calcDiffSafetyB}{0.4917 - 0.4872}
\pgfmathsetmacro{\calcDiffSafetyC}{0.4896 - 0.4872}
\pgfmathsetmacro{\calcDiffSafetyD}{0.4895 - 0.4872}
\pgfmathsetmacro{\calcDiffSafetyE}{0.4891 - 0.4872}
\pgfplotsset{compat=1.18}
\begin{document}
\maketitle

\begin{abstract}
Checkpoint merging is a technique for combining multiple model snapshots into a single superior model, potentially reducing training time for large language models. This paper explores checkpoint merging in the context of parameter-efficient fine-tuning (PEFT), where only small adapter modules (e.g. LoRA) are trained. We propose Metrics-Weighted Averaging (MWA), a simple yet effective method to merge model checkpoints by weighting their parameters according to performance metrics. In particular, we investigate weighting by training loss and by training steps, under the intuition that lower-loss or later-step checkpoints are more valuable. We introduce a formula with a penalty factor to adjust weight distribution, requiring only one hyperparameter regardless of the number of checkpoints. Experiments on three fine-tuning tasks (mathematical reasoning, preference alignment, and general instruction tuning) show that MWA consistently produces merged models that outperform the naive uniform average of checkpoints. Notably, loss-weighted merging often yields the best results, delivering up to ~5\% higher task accuracy than the baseline uniform merge and even surpassing the final individual checkpoint’s performance. These findings validate checkpoint merging for PEFT and demonstrate that a metric-driven weighting heuristic can efficiently boost model performance with minimal computational overhead.
\end{abstract}


\section{Introduction}
Model merging is a technique that merges the parameters of two or more deep learning models into a single model \citep{yang2024modelmergingllmsmllms}. This method has seen great success in research community recently \citep{goddard2025arceesmergekittoolkitmerging}, by merging different models finetuned on multiple datasets and obtaining a single model with all of their capabilities \citep{pmlr-v162-wortsman22a}. In this paper, we explore checkpoint merging, which is akin to model merging and merges the parameters of two or more checkpoints within the same training run into the single model checkpoint. The primary objectives of checkpoint merging is to create 1) a merged checkpoint \textit{post-training} that is superior to all available checkpoints, or 2) a merged checkpoint superior to the model checkpoint at current time step \textit{during training}. The latter objective, which may help to reduce training time, is especially beneficial for state-of-the-art large language models (LLMs) pre-training which requires substantial compute resources to train \citep{hoffmann2022trainingcomputeoptimallargelanguage}.

The problem with existing model merging methods is that they are either overly naive (e.g. Uniform Soup \citep{pmlr-v162-wortsman22a}) or overly complex (e.g. Fisher-Weighted Averaging \citep{matena2022mergingmodelsfisherweightedaveraging}). Simpler methods often underfits and struggle to derive an optimal model soup, while complex methods require high compute or implementation costs in exchange for relatively marginal gains in performance, which is also often not generalizable. Our proposed method, Metrics Weighted Averaging (MWA), strikes a balance by adopting a heuristical approach that is (1) simple enough to be efficiently implemented and, at the same time, (2) yield significant performance gains.

Metrics-Weighted Averaging (MWA) ties the importance of a model to its performance in a given metric relative to other models in the model soup. To that end, we propose various formulas to heuristically calculate the weight of the checkpoints using different metrics. We then performed several experiments on well-studied finetuning datasets and benchmarks to empirically show that our MWA method can bring superior performance while introducing minimal overhead in weight computation. The experiment results demonstrate the ability of Metrics Weighted Averaging to identify an improved distribution of weights in a model soup within a short number of iterations. 

Recently, Parameter-Efficient Finetuning (PEFT) has become a popular alternative to full-finetuning LLMs due to their cost-effectiveness and powerful performance. While traditional model soups approach is applied on fully-tuned model checkpoints, we challenge this scheme and apply our MWA on the PEFT checkpoints. This could unlock even more cost-effective finetuning for higher performance gains on the target task.

In this study, we specifically examine checkpoints consisting of low-rank adaptation (LoRA) modules. LoRA is a PEFT approach that has gained significant traction in recent years \citep{hu_lora_2021}. We show notable performance gains on applying LoRA MWA on these checkpoints, especially for Math datasets: our best performing model merged from checkpoints via LoRA MWA outperformed (1) the last checkpoint in the model soup by 2.62\%, (2) its Uniform Soup version by 2.84\%, and (3) the \sout{last merging checkpoint }final checkpoint in the model soup by 5.84\% in the challenging GSM-Plus math benchmark.

The contributions of this paper can be summarized as follows: 
\begin{enumerate}
    \item We validate the effectiveness of checkpoint merging specifically in the context of parameter-efficient fine-tuning.
    \item We propose a metric-driven approach to find the optimal merging weights and its corresponding formula. 
    \item Lastly, we empirically demonstrate the effectiveness of our proposed method on various finetuning tasks.
\end{enumerate}

In particular, we discover that training loss serves as a highly reliable metric for metrics-weighted averaging.

\section{Related Work}
\paragraph{Model Merging} or model fusion, is a deep learning technique that combines the parameters or predictions of two or more pre-training models into a single model \citep{pmlr-v162-wortsman22a}. It improves overall performance by ensembling the abilities of multiple models to compensate for the biases and errors of a single model. Model merging serves as an efficient technique to create multi-task models, or improve the performance on a single task, without the need for retraining \citep{ilharco2023editing}. The most common type of model merging is parameter-wise merging, where corresponding parameters from multiple models are merged via a linear combination to obtain a new set of parameters. State-of-the-art model merging methods predominantly use parameter-wise merging, and so does our proposed metrics-weighted averaging method.

\paragraph{Checkpoint Merging} is a specific subset of model merging that merges multiple checkpoint models in a single training trajectory. Existing literature on checkpoint merging is severely limited. To the best of our knowledge, there is only one other paper that extensively studies checkpoint merging in the context of transformer-based language models\citep{liu_checkpoint_2024}. Liu et al. explores parameter-wise checkpoint merging via checkpoint merging, specifically in the full pre-training stage. This paper adds on to existing research in threefold. (1) It introduces a new technique for checkpoint merging. (2) It explores checkpoint merging on post-training stages, including supervised fine-tuning and human preference alignment. (3) It studies the effectiveness of checkpoint merging on parameter-efficient modules.

\paragraph{Parameter-Efficient Fine-Tuning (PEFT)} reduces the number of parameters that need to be fine-tuned by updating a small subset of the model's parameters rather than the entire network \citep{xu2023parameterefficientfinetuningmethodspretrained}. This approach significantly reduces computational cost and memory usage while maintaining a similar performance to full fine-tuning. It has seen increasing popularity with the advent of large language models, with LoRa \citep{hu_lora_2021} emerging as the most prominent PEFT method. While many LoRa derivatives such as DoRa \citep{pmlr-v235-liu24bn} and QLoRa \citep{dettmers2023qloraefficientfinetuningquantized} has emerged since LoRa's inception, this paper mainly focuses on the vanilla LoRa implementation. 

\section{Method}

In this seection, we will describe our method, Metrics-Weighted Averaging (MWA) and its two sub-categories, Loss-Weighted Averaging and Steps-Weighted Averaging.

\subsection{Metrics-Weighted Averaging}

Metrics-Weighted Averaging, is a proposed model merging technique which hinges on the belief that by nature, the relative performance of models based on certain metrics can be directly translated to their relative importance, and by extension, their relative weights in a model soup.

In this paper, we focus on checkpoint merging on a single training run. In our experiments, we test two specific metrics, steps and loss. Steps refers to the number of training steps elapsed when a checkpoint is saved while loss refers to the training loss of the checkpoint. We chose these two metrics because (1) they are readily available and thus serve as good starting points for experimentation and (2) heuristically, there is likely correlation between these two metrics and the importance of individual checkpoints.

\paragraph{Loss-Weighted Averaging}

Based on intuition, we assign more weight to a checkpoint with a lower loss. The basic formula expresses how to merge \( n \) checkpoints with losses \( \{ \ell_1, \ell_2, \dots, \ell_n \} \). Let \(\ell_x > 0\) denote the loss associated with \(x\). Define the unnormalized weights \(\phi_x\) as the inverses of the losses:

\begin{equation}
\label{eq:lwa}
\phi_x = \frac{1}{\ell_x}.
\end{equation}

Then, we obtain the normalized weights using:

\begin{equation}
\label{eq:norm}
w_x = \frac{\phi_x}{\sum_{i=1}^{n} \phi_i} = \frac{\dfrac{1}{\ell_x}}{\sum_{i=1}^{n} \dfrac{1}{\ell_i}}
\quad \text{and} \quad 
\sum_{x=1}^{n} w_x = 1.
\end{equation}

However, in Eq.~\ref{eq:lwa}, the distribution of weights relies entirely on the relative difference of the loss, which will be smaller in scale towards the end of training. Hence, this will not be able to bring significant changes compared to uniform model soup. To amplify the weight variation, we modified the formula to include two additional variables: the penalty factor \( p \) and the power factor \( q \). We also include the term \(\text{pos}(\ell_x)\) to represent the zero-indexed position of \(\ell_x\) when the losses are sorted from lowest to highest (best to worst).

\begin{equation}
\label{eq:lwa-final}
\phi_{x|p,q} = p^{q^{\text{pos}(\ell_x)}} \cdot \frac{1}{\ell_x} \quad \text{where} \quad 
q > 1 \text{,} \quad p < 1
\end{equation}

The weights are then normalized using the Eq.~\ref{eq:norm}.

\paragraph{Steps-Weighted Averaging}

For steps-weighted averaging, the objective is in the other direction. Intuitively, a checkpoint trained for a longer period is expected to be more performant, therefore we will assign more weight to a checkpoint saved at a later step. The formula is similar, except that we use the metric value instead of its reciprocal. For the basic formula, $\phi_x = \text{s}_x$ when merging \( n \) checkpoints with steps \( \{ \text{s}_1, \text{s}_2, \dots, \text{s}_n \} \). Again, the basic formula will not suffice for practical application. The enhanced formula can be expressed as follows given penalty factor \( p \), power factor \( q \), and the term \(\text{pos}(\text{s}_x)\) representing the zero-indexed position of \(\text{s}_x\) when the steps are sorted from highest to lowest (best to worst):

\begin{equation} 
\phi_{x|p,q} = p^{q^{\text{pos}(\text{s}_x)}} \cdot \text{s}_x \quad \text{where} \quad 
q > 1 \text{,} \quad p < 1
\end{equation}

\begin{equation}
w_x = \frac{\phi_{x|p,q}}{\displaystyle\sum_{i=1}^{n} \phi_{i|p,q}}
\quad \text{and} \quad 
\sum_{x=1}^{n} w_x = 1
\end{equation}

\paragraph{Penalty and Power Factors}

Penalty factor determines the base level of decay, affects how quickly the weights decrease overall, and modifies the magnitude of weights uniformly across all positions. Power factor determines the steepness of the decay curve, afects how sensitive the weights are to the ranking (position) of the corresponding checkpoint, and modifies the relative differences between consecutive weights. In our experiments, we fix the value of \( q \) heuristically based on the distribution of checkpoint metrics values. We vary the values of \( p \) to produce the optimal set of weights for merging.

\subsection{Dichotomous Abstraction}
Our weighted average formula is intrinsically metric-agnostic, in the sense that it can be generalized to accommodate all types of metrics as long as they fit into one of two primary objective types: "Min" and "Max". "Min" assigns greater weight to lower values, and training loss is an example of a "Min" metric where a smaller loss is more desirable. "Max" assigns greater weight to higher values, and checkpoint steps is an example of a "Max" metricwhere a checkpoint trained for a longer period is expected to perform better.

The beauty of this abstraction lies in its generality: the two weighted average formulae can be adapted across the universe of metrics according to the metric's objective type (Min or Max). This approach is metrics-agnostic because it does not rely on the specific nature of the metric aside from minor variations in the weighting mechanism, for instance, adjusting the power factor to an optimal value based on the metric.

\section{Model Training}

\subsection{Preliminary}

Our paper explores parameter-wise checkpoint merging of LoRa modules. In the experiment, we performed supervised fine-tuning (SFT)\citep{zhang_instruction_2024} on two models and human preference alignment on the remaining model using simple preference optimization (SimPO)\citep{meng_simpo_2024}. These three models are trained on datasets designed for general knowledge, math, and safety alignment respectively.

\subsection{Training Details}
\textbf{Hardware}

The models are trained on a single node of 8x H100 GPUs using Distributed Data Parallelism \citep{bai_modern_2022}.

\textbf{Models}

The models are trained using Gemma-2b \citep{gemmateam2024gemmaopenmodelsbased} as the base model. For the safety alignment model, we trained on the instruct version of Gemma-2b. In total, we trained 3 models, each associated with a specific domain: (1) \textbf{MWA-Math} Gemma-2b completion model supervised fine-tuned on the orca-math \citep{mitra_orca-math_2024} instruct dataset. (2) \textbf{MWA-Align}: Gemma-2b-it instruct model aligned on the hh-rlhf \citep{bai_training_2022} helpfulness and harmlessness alignment dataset using simple preference optimization. (3) \textbf{MWA-Inst}: Gemma-2b completion model supervised fine-tuned on the openhermes 2.5 \citep{OpenHermes2_5} instruct dataset. The training recipe is showin in Table~\ref{tab:trainingrecipe}.

\begin{table}[ht]
\centering
\caption{Training recipe for the 3 experiments}
\setlength{\tabcolsep}{10pt} 
\begin{tabular}{lccc}
  \toprule
  & \textbf{sft\_lora\_math} & \textbf{simpo\_lora\_hh-rlhf} & \textbf{sft\_lora\_openhermes-2\_5} \\
  \midrule
  \textbf{Base Model} & gemma-2b & gemma-2b-it & gemma-2b \\
  \textbf{Block Size} & 2048 & 2048 & 2048 \\
  \textbf{Global Batch Size} & 32 & 16 & 32 \\
  \textbf{Learning Rate} & 2e-5 & 1e-6 & 2e-5 \\
  \textbf{Optimizer} & AdamW & AdamW & AdamW \\
  \textbf{n Epochs} & 3 & 3 & 3 \\
  \textbf{n Steps} & 17814 & 28641 & 89199 \\
  \textbf{n Checkpoints} & 18 & 29 & 23 \\
  \textbf{Checkpoint Intervals} & 1000 & 1000 & 4000 \\
  \bottomrule
\end{tabular}
\label{tab:trainingrecipe}
\end{table}

\section{Experiment}

After model training, we evaluated the checkpoints over a wide range of metrics and tasks. There are three criteria for success: (1) The overarching model training is successful; (2) A k-last metrics-weighted model soup is found to significantly outperform its k-last uniform soup counterpart; (3) A k-last metrics-weighted model soup is found to significantly outperform the last merging checkpoint [A] and marginally outperform the final checkpoint [B].

In the experiments, models capabilities are judged in two ways. Primarily, we used domain-specific evaluation benchmarks to compare the performance of the model. In addition, we also tested the ability of our models to learn the training corpora. We term this ability as "in-distribution" learning, and we evaluate it using the validation loss on a subset of the training corpora.

\subsection{Preliminary}

\paragraph{Terminology}\mbox{}\\

We define key terminologies used in the context of model checkpointing and merging strategies:

\begin{itemize}
    \item \textbf{Final Checkpoint}: The checkpoint saved at the end of the training run.
    
    \item \textbf{Last Merging Checkpoint}: The last (or most recent) checkpoint utilized in the merging process.
    
    \item \textbf{k\_last}: Refers to the process of merging the last \( k \) checkpoints, excluding the final checkpoint.
    
    \item \textbf{Uniform Soup}: A method in existing literature where the weights of the last \( k \) checkpoints are averaged to create a new model. Uniform Soup serves as the baseline in this experiment to evaluate the effectiveness of the proposed method. In mathematical terms, if \(\theta_1, \theta_2, \dots, \theta_k\) represent the weights of the \( k \) selected checkpoints, the weights of the uniform soup model \(\theta_{\text{uniform soup}}\) are computed as:
    
    \[
    \theta_{\text{uniform soup}} = \frac{1}{k} \sum_{i=1}^{k} \theta_i
    \]
\end{itemize}

\paragraph{Naming Conventions for Merged Checkpoints}\mbox{}\\

To easily communicate the nature of the merge, we adopted the following naming convention for the merged checkpoints:

\begin{verbatim}
last_{k}[_{m}]_{unweighted|loss_pf-{n}_{d}|steps_pf-{n}_{d}}
\end{verbatim}

\begin{enumerate}
    \item \texttt{last\_\{k\}}:
    \begin{itemize}
        \item \textbf{\{k\}}: The number of checkpoints in the model soup. \texttt{k=5} merges the 5 latest checkpoints. 
    \end{itemize}
    
    \item \texttt{[\_\{m\}]} (Optional):
    \begin{itemize}
        \item \textbf{\{m\}}: The number of intervals between checkpoints. If \texttt{m=2}, checkpoints are merged every two intervals.
        \item \textbf{Example}: For a sequence \{90, 80, 70, 60, 50, ...\}, \texttt{m=2} merges checkpoints \{90, 70, 50, ...\}.
    \end{itemize}

    \item \texttt{\{unweighted\textbar{}loss\_pf-\{n\}\_\{d\}\textbar{}steps\_pf-\{n\}\_\{d\}\}}:
    \begin{itemize}
        \item \textbf{unweighted | loss | steps }: specifies if the checkpoints are merged using Uniform Soup, Loss-Weighted Averaging, or Steps-Weighted Averaging respectively.
        
        \item \textbf{pf-\{n\}\_\{d\}}: penalty factor applied to the merge, with \{n\} and \{d\} referring to the integer and decimal terms of the penalty factor value respectively.
    \end{itemize}
\end{enumerate}

\paragraph{Merging Steps}\mbox{}\\

\begin{enumerate}
    \item Identify the evaluation metric(s) or benchmark(s) to test based on training objective. Conduct an initial validation to ensure the training run is successful.
    \item Begin with \textit{uniform soup merging} of the last $k$ checkpoints, where $k = \{2, 3, 4, 5, 8, 10\}$. Evaluate the results.
    \item Shortlist the top 1-2 performing models. Conduct \textit{uniform soup merging (intervals)} if $k$ is sufficiently small.
    \item Shortlist the top 1-2 performing models from the \textit{uniform soup-uniform soup (intervals)} union set.
    \item For each shortlisted model, perform loss-weighted and steps-weighted average merging. Vary the penalty factors to get different model soups.
\end{enumerate}

\subsection{Results}

In this section, we will present the benchmark results on well established benchmarks in instruction following, math, and alignment. 

\paragraph{Supervised Fine-Tuning on Math}\mbox{}\\

\begin{figure}[ht]
    \centering
\begin{tikzpicture}
\begin{axis}[
    ylabel=GSM Weighted Average,
    enlargelimits=0.15,
    legend style={at={(0.5,0.2)},
    anchor=north,legend columns=-1},
    ybar stacked,
    ymin=0.2,
    ymax=0.235,
    bar width=20pt,
    xtick=data,
    x dir=reverse,
    ymajorgrids=true,
    grid style=dashed,
    symbolic x coords={last\_10\_loss\_pf-0\_7, last\_4\_3\_loss\_pf-0\_8, last\_10\_steps\_pf-1\_0, last\_4\_3\_loss\_pf-0\_5, last\_10\_steps\_pf-1\_05},
    xtick=data,
    xticklabel style={anchor=east, xshift=10pt, rotate=45, xshift=-5pt},
    x tick label as interval=false,
]

\addplot coordinates {(last\_10\_loss\_pf-0\_7,0.21728) (last\_4\_3\_loss\_pf-0\_8,0.21728) (last\_10\_steps\_pf-1\_0,0.21728) (last\_4\_3\_loss\_pf-0\_5,0.21728) (last\_10\_steps\_pf-1\_05,0.21728)};
\addplot coordinates {(last\_10\_loss\_pf-0\_7,\calcDiffMathA) (last\_4\_3\_loss\_pf-0\_8,\calcDiffMathB) (last\_10\_steps\_pf-1\_0,\calcDiffMathC) (last\_4\_3\_loss\_pf-0\_5,\calcDiffMathD) (last\_10\_steps\_pf-1\_05,\calcDiffMathE)};

\legend{Math (baseline), Math}

\node[anchor=south] at (axis description cs:0.9,0.75) {\textbf{+5.05\%}};

\end{axis}
\end{tikzpicture}
\caption{Merged checkpoints scored on GSM Weighted Average benchmark. The weighted average is calculated with weights \(w_{\text{gsm8k}} = 0.3\) and \(w_{\text{gsmplus}} = 0.7\). Here, Math (baseline) refers to the math weighted-average score of the last merging checkpoint.}
    \label{fig:gsmresult}
\end{figure}
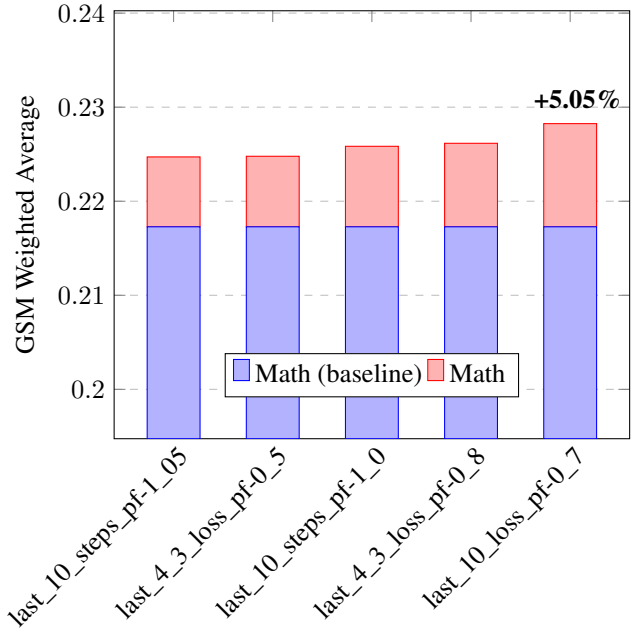

We finetuned our model using LoRA adapters on math datasets (Orca-Math) to obtain a stream of \textbf{MWA-Math} checkpoints. Since the model is fine-tuned to improve mathematical capabilities, we evaluate its performance using math benchmarks GSM8k \citep{cobbe_training_2021} and GSM-Plus \citep{li_gsm-plus_2024}. We then calculate the overall performance using the weighted average, representing math capabilities. As GSM-Plus is comparatively more challenging and robust, we place greater weight on it than GSM8k.

We then begin the checkpoint merging steps, starting from the initial validation of the training run. The results are shown in Figure~\ref{fig:gsmresult}, and more details are available in Table~\ref{tab:gsmresult}. Since the last merging checkpoint and the final checkpoint significantly outperformed the base model (gemma-2b), we can proceed to the next steps. From the uniform soup-uniform soup with intervals union set, we selected the top two performing models, \texttt{last\_10\_unweighted}, and \texttt{last\_4\_3\_unweighted}. We then perform loss-weighted averaging and steps-weighted averaging for each model, varying the distribution of weights by adjusting the penalty factor.

The top five best performing models all happened to be models derived from metrics-weighted averaging, demonstrating its prowess over the naive Uniform Soup approach. The best performing model, \texttt{last\_10\_loss\_pf-0\_7}, improved math capability by 5.05\% and 1.76\% from the last merging checkpoint and final checkpoint respectively. The experiment therefore fulfills all three criteria for success.

\paragraph{Alignment RLHF Training}\mbox{}\\

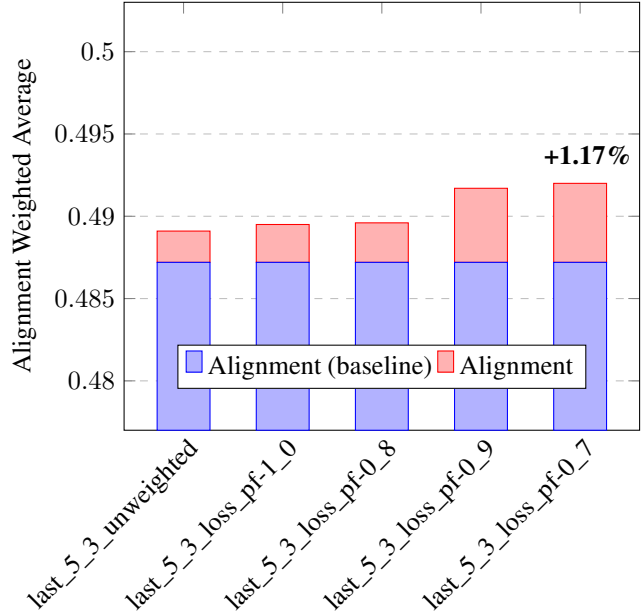
\begin{figure}[ht]
    \centering
\begin{tikzpicture}
\begin{axis}[
    ylabel=Alignment Weighted Average,
    enlargelimits=0.15,
    legend style={at={(0.5,0.2)},
    anchor=north,legend columns=-1},
    ybar stacked,
    ymin=0.48,
    ymax=0.50,
    bar width=20pt,
    xtick=data,
    x dir=reverse,
    ymajorgrids=true,
    grid style=dashed,
    symbolic x coords={last\_5\_3\_loss\_pf-0\_7, last\_5\_3\_loss\_pf-0\_9, last\_5\_3\_loss\_pf-0\_8, last\_5\_3\_loss\_pf-1\_0, last\_5\_3\_unweighted},
    xtick=data,
    xticklabel style={anchor=east,xshift=10pt,rotate=45,xshift=-5pt},
    yticklabel style={/pgf/number format/.cd, fixed, precision=3},
    x tick label as interval=false,
]

\addplot coordinates {(last\_5\_3\_loss\_pf-0\_7,0.4872) (last\_5\_3\_loss\_pf-0\_9,0.4872) (last\_5\_3\_loss\_pf-0\_8,0.4872) (last\_5\_3\_loss\_pf-1\_0,0.4872) (last\_5\_3\_unweighted,0.4872)};
\addplot coordinates {(last\_5\_3\_loss\_pf-0\_7,\calcDiffSafetyA) (last\_5\_3\_loss\_pf-0\_9,\calcDiffSafetyB) (last\_5\_3\_loss\_pf-0\_8,\calcDiffSafetyC) (last\_5\_3\_loss\_pf-1\_0,\calcDiffSafetyD) (last\_5\_3\_unweighted,\calcDiffSafetyE)};

\legend{Alignment (baseline), Alignment}

\node[anchor=south] at (axis description cs:0.9,0.6) {\textbf{+1.17\%}};

\end{axis}
\end{tikzpicture}
\caption{Merged checkpoints scored on Alignment Weighted Average benchmark. The weighted average is calculated with weights \(w_{\text{toxigen}} = 0.5\) and \(w_{\text{truthfulqa\_mc1}} = 0.25\) and \(w_{\text{truthfulqa\_mc2}} = 0.25\). Here, Alignment (baseline) refers to the alignment weighted-average score of the final checkpoint.}
    \label{fig:alignmentresult}
\end{figure}

In this experiment, we finetuned the model for human preference alignment using Reinforcement Learning from Human Feedback (RLHF), specifically using the SimPO method along with LoRA adapters. We call the checkpoints of these approach \textbf{MWA-Align}. We evaluated the model's performance using alignment benchmarks ToxiGen \citep{hartvigsen_toxigen_2022} and TruthfulQA \citep{lin_truthfulqa_2022}. We then calculate the overall performance using the weighted average, representing alignment capabilities. In the experiment, we assign equal weighting to ToxiGen and TruthfulQA.

We then follow the same steps as in the previous experiment. This time, four out of five of the top five models are metrics-weighted, as shown in Figure~\ref{fig:alignmentresult}. Once again, the results demonstrate the prowess of our method over Uniform Soup. The results also demonstrate the prowess of loss-weighted averaging over steps-weighted averaging, as all the metrics-weighted models in the top five are loss-weighted. The best performing model, \texttt{last\_5\_3\_loss\_pf-0\_7}, improved alignment capability by 1.17\% and 0.99\% from the last merging checkpoint and final checkpoint respectively. Once again, our experiment demonstrated the ability of our proposed method in fulfilling the 3 criteria of success for checkpoint merging.

\subsection{In-Distribution Learning Results}
\paragraph{Supervised Fine-Tuning: sft\_lora\_openhermes-2\_5}\mbox{}\\
\begin{table}[H]
\centering
\begin{minipage}[t]{0.35\textwidth}
    This experiment tests the hypothesis that our method can improve in-distribution learning from the training dataset. The evaluation metric is the validation loss on a subset of the training dataset (i.e. OpenHermes 2.5). \\

    As the objective is in-distribution learning within the training dataset, we skip the initial validation step of comparing against the base model. After following the remaining steps, we end up with \texttt{last\_10\_loss\_pf-0\_75} as the best performing model. By strategically varying the penalty factor values, we are able to identify a loss-weighted model that outperforms its Uniform-Soup equivalent (\texttt{last\_10\_unweighted}) within four steps.
\end{minipage}%
\hfill
\begin{minipage}[t]{0.65\textwidth}
    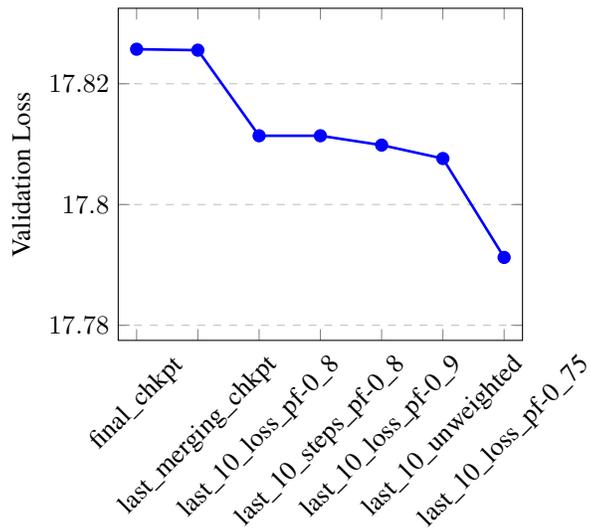
\begin{figure}[H]
    \centering
\begin{tikzpicture}
\begin{axis}[
    ylabel=Validation Loss,
    height=6cm,
    enlargelimits=0.05,
    ymin=17.78,
    ymax=17.83,
    xtick=data,
    x dir=normal,
    ymajorgrids=true,
    grid style=dashed,
    symbolic x coords={
        final\_chkpt,
        last\_merging\_chkpt,
        last\_10\_loss\_pf-0\_8,
        last\_10\_steps\_pf-0\_8,
        last\_10\_loss\_pf-0\_9,
        last\_10\_unweighted,
        last\_10\_loss\_pf-0\_75
    },
    xticklabel style={rotate=45},
    yticklabel style={/pgf/number format/.cd, fixed, precision=3},
    x tick label as interval=false,
]

\addplot[
    mark=*,
    color=blue,
    line width=1pt
] coordinates {
    (final\_chkpt,17.82572)
    (last\_merging\_chkpt,17.82556)
    (last\_10\_loss\_pf-0\_8,17.81139)
    (last\_10\_steps\_pf-0\_8,17.81139)
    (last\_10\_loss\_pf-0\_9,17.80984)
    (last\_10\_unweighted,17.80762)
    (last\_10\_loss\_pf-0\_75,17.79124)
};


\end{axis}
\end{tikzpicture}
\caption{Merged checkpoints scored on OpenHermes-2.5 validation loss}
    \label{fig:my_plot}
\end{figure}
\end{minipage}
\end{table}

\subsection{Comparative Study}
We conducted experiments with existing merging methods, which we shall refer to as "baseline methods". We selected three baseline methods, namely Ties \citep{yadav_ties-merging_2023}, Dare-Ties, and SLERP. The objective of these experiments are twofold: (1) Serve as baselines to compare against the performance of our Metrics-Weighted Averaging method in order to validate the latter's effectiveness and (2) Assess the performance and feasibility of using baseline methods for Parameter-Efficient Checkpoint Merging. 

For Ties and Dare-Ties, we merged the last 3, 5, and 10 checkpoints for sft\_lora\_math, simpo\_lora\_hh-rlhf, and sft\_lora\_openhermes-2\_5 experiments. We included an additional merge only for simpo\_lora\_hh-rlhf for each method, where we merged the last 5 checkpoints with intervals of 3. For SLERP, we conducted pairwise merging for the three experiments. In addition, we also tested using different base models for each merge (first checkpoint vs last checkpoint in the training run). In total, we produced \(\left(2 \times 3 \times 3 + 2 \times 1 + 1 \times 3\right) \times 2 = 46\) baseline merges.

\paragraph{Baseline Results}\mbox{} \\
The baseline methods demonstrate decent improvements from the last merging and final checkpoints from the sft\_lora\_math experiement. The best performing model, dare\_ties\_last\_10\_base\_first, outperformed the final checkpoint and the last merging checkpoint on the GSM Weighted Average benchmark by 1.86\% and 2.72\% respectively. On the simpo\_lora\_hh-rlhf experiment, all three baseline methods underperformed both the final and last merging checkpoints on the Alignment Weighted Average benchmark. Relative to the final checkpoint, the drop in scores ranged from -0.32\% to as much as -2.13\%.

On the sft\_lora\_openhermes-2\_5 experiment, the results are promising, as most merged models created from the three baseline methods are able to lower the validation loss on OpenHermes-2.5 as compared to the final and last merging checkpoints. Notably, the dare\_ties\_last\_10\_base\_last significantly lowered the validation loss from 17.4189 to 17.068.

For full details, refer to Table~\ref{tab:gsmresultbaseline}, Table~\ref{tab:alignmentresultbaseline}, and Table~\ref{tab:validationlossbaseline} for sft\_lora\_math, simpo\_lora\_hh-rlhf, and sft\_lora\_openhermes-2\_5 respectively.

\paragraph{Comparison and Analysis}\mbox{} \\

While the raw results from the baseline runs exceed those from the metrics-weighted average runs, it is noteworthy to mention that the performance of the checkpoints also showed consistent improvements from the metrics-weighted average runs. The discrepancies can be explained by the fact that the two runs used different underlying hardware accelerators, and previous studies have shown that different GPUs may produce different results. This means that despite using identical configurations, the inference results, and by extension evaluation performances, are different. To account for the differences, we compare the relative results to the final and last merging checkpoint results of the respective runs.

\begin{figure}[ht]
    \centering
    \begin{subfigure}{0.48\textwidth}
        \centering
        \begin{tikzpicture}
        \begin{axis}[
            ylabel=Relative Improvement (GSM),
            enlargelimits=0.05,
            legend style={at={(0.5,1.25)}, anchor=north, legend columns=2, font=\small},
            ybar stacked,
            ymin=0.28,
            ymax=6,
            bar width=6pt,
            xtick=data,
            x dir=reverse,
            ymajorgrids=true,
            grid style=dashed,
            symbolic x coords={top\_1, top\_2, top\_3, top\_4, top\_5},
            xticklabel style={rotate=45, font=\footnotesize},
            y tick label style={font=\footnotesize},
            yticklabel=\pgfmathprintnumber{\tick}\,\%,
            width=\textwidth,
            height=0.8\textwidth,
        ]
        
        \addplot coordinates {(top\_1,2.72133202) (top\_2,2.555724644) (top\_3,2.461731268) (top\_4,2.211082267) (top\_5,1.861964014)};
        
        \addplot coordinates {(top\_1,2.32866798) (top\_2,1.534275356) (top\_3,1.478268732) (top\_4,1.248917733) (top\_5,1.558035986)};
        
        \legend{Baseline, Metrics-Weighted Averaging}
        
        \end{axis}
        \end{tikzpicture}
    \end{subfigure}
    \hfill
    \begin{subfigure}{0.48\textwidth}
        \centering
        \begin{tikzpicture}
        \begin{axis}[
            ylabel=Decrease in Validation Loss,
            legend style={at={(0.5,1.25)}, anchor=north, legend columns=2, font=\small},
            enlargelimits=0.05,
            ymin=-2.1,
            ymax=0,
            xtick=data,
            x dir=normal,
            ymajorgrids=true,
            grid style=dashed,
            symbolic x coords={top\_5, top\_4, top\_3, top\_2, top\_1},
            xticklabel style={rotate=45, font=\footnotesize},
            yticklabel=\pgfmathprintnumber{\tick}\,\%,
            width=\textwidth,
            height=0.8\textwidth,
        ]
        
        \addplot[
            mark=square*,
            color=red,
            line width=1pt
        ] coordinates {
            (top\_5,-0.11654008)
            (top\_4,-0.2198761116)
            (top\_3,-0.2738404836)
            (top\_2,-1.151622663)
            (top\_1,-2.014478526)
        };
        
        \addplot[
            mark=triangle*,
            color=green,
            line width=1pt
        ] coordinates {
            (top\_5,-0.07949259378)
            (top\_4,-0.07949259378)
            (top\_3,-0.08818797278)
            (top\_2,-0.1006419995)
            (top\_1,-0.1925325207)
        };
        
        \legend{Baseline, Metrics Weighted Averaging}
        
        \end{axis}
        \end{tikzpicture}
    \end{subfigure}
    \caption{Comparison of Baseline and Metrics-Weighted Averaging methods relative to their respective last merging checkpoints. Left: \% increase in GSM Weighted Average benchmark. Right: \% decrease in validation loss on OpenHermes-2.5.}
    \label{fig:baselinecompare}
\end{figure}
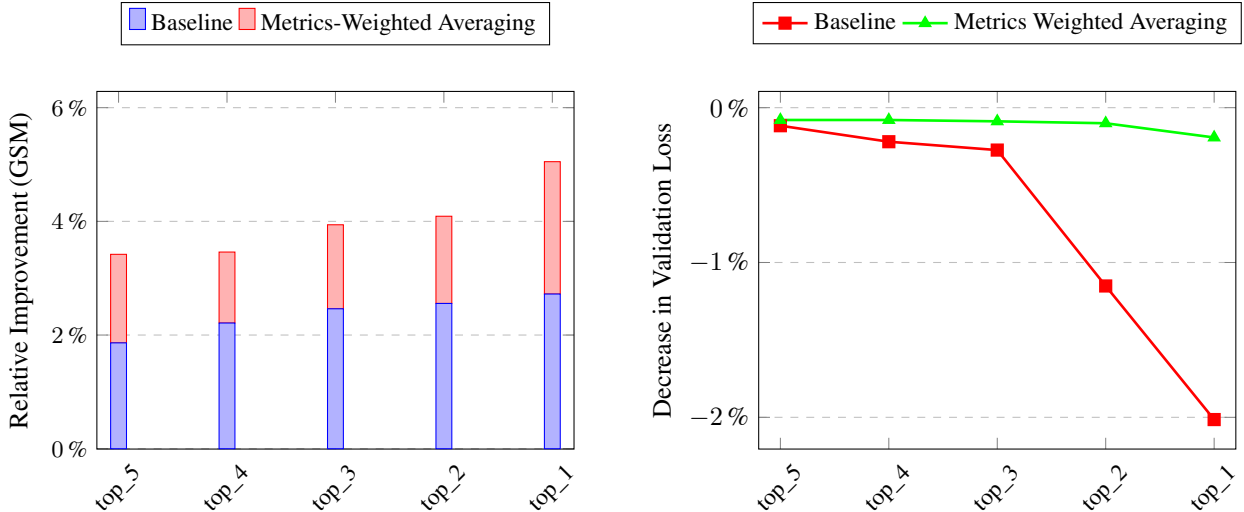

Based on relative results, our metrics-weighted averaging method is more effective compared to all three baseline methods on domain-specific evaluation benchmarks. However, the performance on in-distribution learning produced similar results for both runs, with the latter significantly outperforming in some instances. Figure~\ref{fig:baselinecompare} compares the performance between the top 5 models in each training run for the sft\_lora\_math (domain-specific evaluation benchmark) and sft\_lora\_openhermes-2\_5 (in-distribution learning) experiments. Note that simpo\_lora\_hh-rlhf is excluded from the figure, as the differences are too pronounced to require illustration.

\subsection{Key Findings}
\paragraph{Metrics-Weighted Averaging generally beat existing merging methods.}\mbox{}\\ 
In the study, we evaluated the performance of metrics-weighted averaging against several methods in existing literature. In the first run, we demonstrated that metrics-weighted averaging can improve the results of the naive uniform soup approach by a significant margin. In the second run, we evaluated three baseline methods, Ties, Dare-Ties, and SLERP, and demonstrated the superiority of metrics-weighted averaging in terms of improving domain-specific benchmarks results. However, the baseline methods yield promising in-distribution learning results superior to metrics-weighted averaging.

\paragraph{Experiment results challenge prevailing assumptions about checkpoint merging.}\mbox{}\\
In a previous study on checkpoint merging \citep{liu_checkpoint_2024}, the pilot experiment revealed two key insights: (1) Merging three or four checkpoints offered no significant performance advantage over pairwise merging, and (2) Merging checkpoints that are widely spaced apart often led to a significant drop in performance.

In contrast, our findings challenge these conclusions. Our best performing models were created by merging a range of 4 to 10 checkpoints, with pairwise merging delivering lackluster results in comparison. We also discovered that interval merging, which increases the spacing between checkpoints, actually produced better outcomes than merging adjacent checkpoints in some experiments.

\paragraph{The usefulness of metrics-weighted averaging scales exponentially with number of checkpoints.}\mbox{}\\
Intuitively, the search space for an optimal set of weights increases exponentially with the number of checkpoints in a checkpoint soup. In our proposed weighted average formula, we only tune a single parameter, the penalty factor, irrespective of the size of the checkpoint soup. (For simplicity sake, we suggest fixing the power factor). In the case of pairwise merging, which is represented by $\theta_{\text{new}} = \lambda \cdot \theta_1 + (1 - \lambda) \cdot \theta_2$, we are essentially tuning a single parameter as well. Therefore, the benefits of metrics-weighted averaging is marginal vis-a-vis a randomized hyperparameter tuning approach. However, as the checkpoint soup becomes larger, so does the number of parameters to tune, which results in an exponential growth in the size of the search space. Mathematically, when there are k checkpoints in the checkpoint soup, we need to optimize for k-1 parameters (weights). (The last weight can simply be derived via the condition that sum of all weights equals one). As k becomes large, our metrics-weighted method significantly reduces the search space.

Empirically, we observed that metrics-weighted averaging for pairwise merges marginally outperforms the Uniform Soup equivalent at best, and often underperforms the latter. Going through the same number of steps, a merge size of 5 - 10 checkpoints yields much better results.

\paragraph{Some metrics are more equal than others.}\mbox{}\\
Loss-weighted averaging consistently yield superior model soups compared to steps-weighted averaging. This disparity is most striking in the safety alignment experiment, where four out of the top five metrics-weighted models are loss-weighted. This highlights that the effectiveness of metrics-weighted averaging is heavily influenced by the nature and quality of the metrics used. Not all metrics will lead to optimal results, and choosing the correct metric is an ongoing research question.

\section {Scope and Limitations}
\subsection{Scope}
The study focuses on developing and testing a proposed model merging method, metrics-weighted averaging. The primary objective is to explore the feasibility of this approach and evaluate the effectiveness of its corresponding formula compared to the baseline method, Uniform Soup. The study specifically targets checkpoint merging on LoRa parameter-efficient modules derived from LLM post-training, particularly in the contexts of supervised fine-tuning and human preference alignment.

\subsection{Limitations}

While the study demonstrates promising results, several limitations must be acknowledged. Due to constraints on computational resources, our method was not evaluated on models trained with full fine-tuning or pre-training, which could further assess its effectiveness. In addition, the study was limited to only two types of metrics. Although training loss has been demonstrated as an effective metric for weighted averaging, experimenting with additional metrics could better overview of the relationship between metrics and their effectiveness. Furthermore, our study focuses specifically on checkpoint merging within a single training run aimed at improving performance on a single task. A more robust evaluation could involve applying our method to a generalized model merging approach, such as multi-task learning, to better assess its broader applicability and effectiveness.

\section {Conclusion}

In this paper, we introduce **Metrics-Weighted Averaging (MWA)**, a simple yet powerful method for checkpoint merging in parameter-efficient fine-tuning (PEFT) scenarios. Our approach assigns weights to checkpoints based on performance metrics such as training loss and training steps, with the goal of improving model quality through smarter merging. By incorporating a penalty and power factor into our weighting scheme, MWA offers a flexible, tunable framework that requires only a single hyperparameter regardless of the number of checkpoints.

Through extensive experiments on three domains: mathematical reasoning, alignment via RLHF, and instruction tuning, we demonstrate that MWA consistently outperforms naive uniform averaging. In particular, **loss-weighted averaging** emerges as the most effective merging strategy, leading to improvements of up to \textbf{5.05\%} in task-specific benchmarks and even surpassing the performance of final model checkpoints. Comparative analysis with state-of-the-art baseline merging methods further confirms the strength of MWA, particularly in domain-specific generalization.

In addition, we show that the effectiveness of MWA scales well with the number of checkpoints involved and that merging spaced apart checkpoints can lead to significant gains, challenging prior assumptions in the literature. Although our experiments focus on LoRA modules within a single training run, the results underscore the broader potential of metric-driven merging strategies in low-resource, high-efficiency model development pipelines.

Ultimately, our findings validate that **checkpoint merging in PEFT is not only feasible but also highly beneficial** when guided by the right metric heuristics. We hope this work inspires further research into generalized metric-based merging techniques and their applications in broader settings such as multitask and multirun model fusion.

\bibliographystyle{unsrtnat}
\bibliography{references}  

\appendix
\section{Further Details on Evaluation Results}

\begin{table}[ht]
\centering
\caption{Model Performance on GSM8K and GSM Plus Benchmarks}
\begin{threeparttable}
\begin{tabular}{lccc}
\toprule
\textbf{Model} & \textbf{GSM8K} & \textbf{GSM Plus} & \textbf{Weighted Average\tnote{*}} \\
\midrule
last\_10\_loss\_pf-0\_7 & 0.304  & 0.1958 & 0.22826 \\
& & & \textbf{+5.05\%} / \textbf{+1.76\%}\tnote{+} \\
last\_4\_3\_loss\_pf-0\_8 & 0.3017 & 0.1938 & 0.22617 \\
& & & \textbf{+4.09\%} / \textbf{+0.83\%} \\ 
last\_10\_steps\_pf-1\_0 & 0.3055 & 0.1917 & 0.22584 \\
& & & \textbf{+3.94\%} / \textbf{+0.68\%} \\ 
last\_4\_3\_loss\_pf-0\_5 & 0.2934 & 0.1954 & 0.2248 \\
& & & \textbf{+3.46\%} / \textbf{+0.22\%} \\ 
last\_10\_steps\_pf-1\_05 & 0.2980 & 0.1933 & 0.22471 \\
& & & \textbf{+3.42\%} / \textbf{+0.18\%} \\ 
\midrule
last\_merging\_checkpoint & 0.2926 & 0.1850 & 0.21728 \\
& & & \textbf{+76.9\%}\tnote{\textasciicircum} \\ 
final\_checkpoint & 0.3025 & 0.1908 & 0.22431 \\
& & & \textbf{+82.6\%} \\ 
\midrule
gemma-2b & 0.1751 & 0.1004 & 0.12281 \\
\bottomrule
\end{tabular}
\begin{tablenotes}
\small
\item[*] The weighted average is calculated with weights \(w_{\text{gsm8k}} = 0.3\) and \(w_{\text{gsmplus}} = 0.7\). 
\item[+] The values indicate the percentage increase relative to the last merging checkpoint and the final checkpoint, respectively.
\item[\textasciicircum] The values indicate the percentage increase relative to the base model.
\end{tablenotes}
\end{threeparttable}
\label{tab:gsmresult}
\end{table}

\begin{table}[ht]
\centering
\caption{Model Performance on ToxiGen and TruthfulQA Benchmarks}
\begin{threeparttable}
\begin{tabular}{lcccc}
\toprule
\textbf{Model} & \textbf{ToxiGen} & \textbf{TruthfulQA MC1 / MC2} & \textbf{Weighted Average\tnote{*}} \\
\midrule
last\_5\_3\_loss\_pf-0\_7 & 0.5543 & 0.3452 / 0.5142 & 0.4920 \\
& & & \textbf{+1.17\%} / \textbf{+0.99\%}\tnote{+} \\
last\_5\_3\_loss\_pf-0\_9 & 0.5543 & 0.3439 / 0.5143 & 0.4917 \\
& & & \textbf{+1.11\%} / \textbf{+0.92\%} \\
last\_5\_3\_loss\_pf-0\_8 & 0.5543 & 0.3366 / 0.5132 & 0.4896 \\
& & & \textbf{+0.68\%} / \textbf{+0.49\%} \\
last\_5\_3\_loss\_pf-1\_0 & 0.5553 & 0.3354 / 0.5121 & 0.4895 \\
& & & \textbf{+0.66\%} / \textbf{+0.48\%} \\
last\_5\_3\_unweighted & 0.5553 & 0.3354 / 0.5105 & 0.4891 \\
& & & \textbf{+0.58\%} / \textbf{+0.40\%} \\
\midrule
last\_merging\_checkpoint & 0.5553 & 0.3268 / 0.5114 & 0.4872 \\
& & & \textbf{+9.9\%}\tnote{\textasciicircum} \\ 

final\_checkpoint & 0.5553 & 0.3244 / 0.5102 & 0.4863 \\
& & & \textbf{+9.7\%} \\ 

\midrule
gemma-2b-it & 0.5128 & 0.2901 / 0.4583 & 0.4435 \\
\bottomrule
\end{tabular}
\begin{tablenotes}
\small
\item[*] The weighted average is calculated with weights \(w_{\text{toxigen}} = 0.5\) and \(w_{\text{truthfulqa\_mc1}} = 0.25\) and \(w_{\text{truthfulqa\_mc2}} = 0.25\). 
\item[+] The values indicate the percentage increase relative to the last merging checkpoint and the final checkpoint, respectively.
\item[\textasciicircum] The values indicate the percentage increase relative to the base model.
\end{tablenotes}
\end{threeparttable}
\label{tab:alignmentresult}
\end{table}

\begin{table}[ht]
\centering
\caption{Model Performance on GSM8K and GSM Plus Benchmarks (Baseline Methods)}
\begin{threeparttable}
\begin{tabular}{lccc}
\toprule
\textbf{Model} & \textbf{GSM8K} & \textbf{GSM Plus} & \textbf{Weighted Average\tnote{*}} \\
\midrule
dare\_ties\_last\_3\_base\_first & 0.3177 & 0.1917 & 0.2295 \\
& & & \textbf{+1.86\%} / \textbf{+2.72\%}\tnote{+} \\
ties\_last\_5\_base\_last & 0.3146 & 0.1925 & 0.2291 \\
& & & \textbf{+1.70\%} / \textbf{+2.56\%} \\
dare\_ties\_last\_5\_base\_last & 0.3139 & 0.1925 & 0.2289 \\
& & & \textbf{+1.61\%} / \textbf{+2.46\%} \\
ties\_last\_10\_base\_first & 0.3139 & 0.1917 & 0.2284 \\
& & & \textbf{+1.36\%} / \textbf{+2.21\%} \\
ties\_last\_3\_base\_last & 0.3260 & 0.1854 & 0.2276 \\
& & & \textbf{+1.01\%} / \textbf{+1.86\%} \\
slerp\_last\_2\_base\_last & 0.3169 & 0.1871 & 0.2260 \\
& & & \textbf{+0.33\%} / \textbf{+1.17\%} \\
dare\_ties\_last\_5\_base\_first & 0.3207 & 0.1850 & 0.2257 \\
& & & \textbf{+0.18\%} / \textbf{+1.02\%} \\
slerp\_last\_2\_base\_first & 0.3108 & 0.1867 & 0.2239 \\
& & & \textbf{-0.61\%} / \textbf{+0.23\%} \\
ties\_last\_3\_base\_first & 0.3139 & 0.1850 & 0.2237 \\
& & & \textbf{-0.72\%} / \textbf{+0.11\%} \\
ties\_last\_10\_base\_last & 0.3078 & 0.1867 & 0.2230 \\
& & & \textbf{-1.01\%} / \textbf{-0.17\%} \\
ties\_last\_5\_base\_first & 0.3063 & 0.1871 & 0.2229 \\
& & & \textbf{-1.08\%} / \textbf{-0.25\%} \\
dare\_ties\_last\_10\_base\_first & 0.3002 & 0.1892 & 0.2225 \\
& & & \textbf{-1.24\%} / \textbf{-0.41\%} \\
dare\_ties\_last\_3\_base\_last & 0.3108 & 0.1842 & 0.2222 \\
& & & \textbf{-1.38\%} / \textbf{-0.55\%} \\
dare\_ties\_last\_10\_base\_last & 0.2987 & 0.1796 & 0.2153 \\
& & & \textbf{-4.43\%} / \textbf{-3.62\%} \\
\midrule
final\_checkpoint & 0.3184 & 0.1854 & 0.2253 \\
& & & \textbf{+80.1\%}\tnote{\textasciicircum} \\
last\_merging\_checkpoint & 0.3161 & 0.1837 & 0.2234 \\
& & & \textbf{+78.6\%} \\
gemma-2b & 0.1827 & 0.1004 & 0.1251 \\
\bottomrule
\end{tabular}
\begin{tablenotes}
\small
\item[*] The weighted average is calculated with weights \(w_{\text{gsm8k}} = 0.3\) and \(w_{\text{gsmplus}} = 0.7\). 
\item[+] The values indicate the percentage increase relative to the last merging checkpoint and the final checkpoint, respectively.
\item[\textasciicircum] The values indicate the percentage increase relative to the base model.
\end{tablenotes}
\end{threeparttable}
\label{tab:gsmresultbaseline}
\end{table}

\begin{table}[ht]
\centering
\caption{Model Performance on ToxiGen and TruthfulQA Benchmarks (Baseline Methods)}
\begin{threeparttable}
\begin{tabular}{lccc}
\toprule
\textbf{Model} & \textbf{ToxiGen} & \textbf{TruthfulQA MC1 / MC2} & \textbf{Weighted Average\tnote{*}} \\
\midrule
dare\_ties\_last\_5\_base\_first & 0.5543 & 0.3329 / 0.5113 & 0.4882 \\
& & & \textbf{-0.32\%} / \textbf{-0.17\%}\tnote{+} \\
ties\_last\_5\_base\_first & 0.5564 & 0.3317 / 0.5059 & 0.4876 \\
& & & \textbf{-0.44\%} / \textbf{-0.30\%} \\
dare\_ties\_last\_3\_base\_first & 0.5553 & 0.3317 / 0.5065 & 0.4872 \\
& & & \textbf{-0.52\%} / \textbf{-0.38\%} \\
slerp\_last\_2\_base\_first & 0.5553 & 0.3317 / 0.5058 & 0.4870 \\
& & & \textbf{-0.56\%} / \textbf{-0.41\%} \\
ties\_last\_3\_base\_last & 0.5564 & 0.3293 / 0.5051 & 0.4868 \\
& & & \textbf{-0.60\%} / \textbf{-0.46\%} \\
dare\_ties\_last\_3\_base\_last & 0.5543 & 0.3293 / 0.5061 & 0.4860 \\
& & & \textbf{-0.77\%} / \textbf{-0.62\%} \\
ties\_last\_10\_base\_first & 0.5553 & 0.3293 / 0.5038 & 0.4859 \\
& & & \textbf{-0.78\%} / \textbf{-0.64\%} \\
slerp\_last\_2\_base\_last & 0.5543 & 0.3305 / 0.5045 & 0.4859 \\
& & & \textbf{-0.79\%} / \textbf{-0.64\%} \\
ties\_last\_3\_base\_first & 0.5543 & 0.3293 / 0.5053 & 0.4858 \\
& & & \textbf{-0.81\%} / \textbf{-0.66\%} \\
ties\_last\_10\_base\_last & 0.5553 & 0.3317 / 0.5007 & 0.4858 \\
& & & \textbf{-0.82\%} / \textbf{-0.67\%} \\
ties\_last\_5\_base\_last & 0.5543 & 0.3305 / 0.5029 & 0.4855 \\
& & & \textbf{-0.87\%} / \textbf{-0.73\%} \\
ties\_last\_5\_3\_base\_first & 0.5543 & 0.3293 / 0.5032 & 0.4853 \\
& & & \textbf{-0.91\%} / \textbf{-0.77\%} \\
dare\_ties\_last\_5\_3\_base\_last & 0.5564 & 0.3268 / 0.5012 & 0.4852 \\
& & & \textbf{-0.93\%} / \textbf{-0.79\%} \\
dare\_ties\_last\_5\_base\_last & 0.5543 & 0.3293 / 0.5023 & 0.4851 \\
& & & \textbf{-0.96\%} / \textbf{-0.82\%} \\
ties\_last\_5\_3\_base\_last & 0.5553 & 0.3293 / 0.4991 & 0.4848 \\
& & & \textbf{-1.02\%} / \textbf{-0.88\%} \\
dare\_ties\_last\_5\_3\_base\_first & 0.5574 & 0.3097 / 0.5110 & 0.4839 \\
& & & \textbf{-1.20\%} / \textbf{-1.06\%} \\
dare\_ties\_last\_10\_base\_last & 0.5532 & 0.3219 / 0.5030 & 0.4828 \\
& & & \textbf{-1.41\%} / \textbf{-1.27\%} \\
dare\_ties\_last\_10\_base\_first & 0.5532 & 0.3219 / 0.4889 & 0.4793 \\
& & & \textbf{-2.13\%} / \textbf{-1.99\%} \\

\midrule
final\_checkpoint & 0.5564 & 0.3329 / 0.5133 & 0.4898 \\
& & & \textbf{+10.3\%}\tnote{\textasciicircum} \\
last\_merging\_checkpoint & 0.5564 & 0.3305 / 0.5129 & 0.4891 \\
& & & \textbf{+10.1\%} \\

\midrule
gemma-2b-it & 0.5117 & 0.2938 / 0.4590 & 0.4441 \\
\bottomrule
\end{tabular}
\begin{tablenotes}
\small
\item[*] The weighted average is calculated with weights \(w_{\text{toxigen}} = 0.5\), \(w_{\text{truthfulqa\_mc1}} = 0.25\), and \(w_{\text{truthfulqa\_mc2}} = 0.25\).
\item[+] Values indicate the percentage increase relative to the final checkpoint and last merging checkpoint.
\item[\textasciicircum] The values indicate the percentage increase relative to the base model.
\end{tablenotes}
\end{threeparttable}
\label{tab:alignmentresultbaseline}
\end{table}

\begin{table}[ht]
\centering
\caption{Validation Loss (Baseline Methods)}
\begin{threeparttable}
\begin{tabular}{lc}
\toprule
\textbf{Model} & \textbf{Validation Loss} \\
\midrule
dare\_ties\_last\_10\_base\_first & 17.68 \\
dare\_ties\_last\_5\_base\_first & 17.4761 \\
ties\_last\_3\_base\_last & 17.4391 \\
dare\_ties\_last\_3\_base\_last & 17.437 \\
ties\_last\_5\_base\_first & 17.4328 \\
slerp\_last\_2\_base\_first & \textbf{17.4172\tnote{*}} \\
slerp\_last\_2\_base\_last & 17.4137 \\
ties\_last\_3\_base\_first & 17.4135 \\
dare\_ties\_last\_3\_base\_first & 17.3991 \\
ties\_last\_10\_base\_first & 17.3986 \\
ties\_last\_5\_base\_last & 17.3806 \\
dare\_ties\_last\_5\_base\_last & 17.3712 \\
ties\_last\_10\_base\_last & 17.2183 \\
dare\_ties\_last\_10\_base\_last & 17.068 \\

\midrule
final\_checkpoint & 17.4189 \\
last\_merging\_checkpoint & 17.4189 \\
\bottomrule
\end{tabular}
\begin{tablenotes}
\small
\item[*] All models at or below the highlighted model show validation losses lower than those of both the final checkpoint and the last merging checkpoint.
\end{tablenotes}
\end{threeparttable}
\label{tab:validationlossbaseline}
\end{table}







\end{document}